# hauWE: Hausa Words Embedding for Natural Language Processing

Idris Abdulmumin
*Department of Computer Science*
*Ahmadu Bello University, Zaria*

Bashir Shehu Galadanci
*Department of Software Engineering*
*Bayero University, Kano*

*Abstract* – Words embedding (distributed word vector representations) have become an essential component of many natural language processing (NLP) tasks such as machine translation, sentiment analysis, word analogy, named entity recognition and word similarity. Despite this, the only work that provides word vectors for Hausa language is that of [1] trained using fastText, consisting of only a few words vectors. This work presents words embedding models using Word2Vec's Continuous Bag of Words (CBoW) and Skip Gram (SG) models. The models, hauWE (Hausa Words Embedding), are bigger and better than the only previous model, making them more useful in NLP tasks. To compare the models, they were used to predict the 10 most similar words to 30 randomly selected Hausa words. hauWE CBoW's 88.7% and hauWE SG's 79.3% prediction accuracy greatly outperformed [1]'s 22.3%.

## I. INTRODUCTION

Words embedding (distributed word vector representations) have become an essential component of many natural language processing (NLP) tasks such as machine translation [2, 3], sentiment analysis [4] and text classification [5].

To improve the accuracy of low resource machine translation, [6] used words vectors to improve the alignments between words. The authors argued that words of similar meanings should have similar translations and that not all words that are substitutable are synonyms.

Natural Language Processing has in recent years receive a lot of attention from researchers. Despite this, only a few works considered low resource Hausa language especially in the field of word vector representations. Most of the attention is on high resource languages such as English, French, Arabic, Chinese, etc. To the best of our knowledge, the only publicly available dataset for machine translation is the Tanzil dataset, consisting of a meagre 127k parallel sentences. The dataset is also made of less than 10k unique Hausa sentences. This dataset also does not represent the various dialects of Hausa language, which according to [7], include Eastern Hausa (e.g. Kano), Western Hausa (e.g. Sokoto) and dialects from Niger (e.g. Aderanci).

The only work that provides word vectors for Hausa language is that of [1]. The work provides trained models for 294 languages including Hausa. The vectors were trained using fastText [8] and the Hausa vectors model consists of a vocabulary of 4347 words only with more than 1500 English words, representing about 40%.

Learning word representations require a lot of data. The performance of the CBoW and SG varies with the amount of data available. Mikolov et al. [9] have found that with more dataset, CBoW model outperforms SG model. While both models perform well with abundant quantities of training data, the SG model provides good representation for rare words in low resource settings.

In this paper, the work presented describes the processes involved in generating hauWE, an open-source distributed words vector representation (words embedding) consisting of a larger vocabulary to provide the research community with a better model for Hausa NLP tasks. The model is built based on text resources crawled from Hausa news sites on the World Wide Web and is trained based on the gensim's Word2Vec implementation of the CBoW and SG models. The model is created using 692k text documents (sentences) consisting of 18.1 million words – 77 times bigger than the old model – and a vocabulary size of 90k words – 20 times bigger than the old model – among which are about 8k English words – mostly names of persons, cities, etc. – representing just 9% of the entire corpus.

|  | Bojanowski | hauWE |
|---|---|---|
| Total no. of words | 234, 779 | 18, 182, 511 |
| No. of English words | 50, 255 (21%) | 1, 278, 069 (7%) |
| Vocabulary size | 4, 347 | 90, 451 |
| English words | 1, 661 (38%) | 8, 584 (9%) |

*Table 1. Words distribution*

Data generation, data cleaning and preprocessing and choice of hyper-parameters are further explained in the paper. The models are compared with that of [1] – labelled Bojanowski. They were used to predict 10 most similar words to each word in a given set of 30 common words and people names. The two newly created models greatly outperformed the previous model.

**Words Embedding**

Words embedding is a way of representing words as vectors in $\mathbb{R}^n$ where words that share a degree of semantic and syntactic similarity are represented by similar (or closer) vectors. A word can be represented in hundreds of dimensions. The idea of words embedding is building a neural language model [9] that can predict the next word given a set of context vectors [10].

The widely used methods to learn word vector models are [9]'s Continuous Bag of Words (CBoW) and Continuous Skip-gram (SG) models. The CBoW model predicts the most probable word given a context vector while the SG model, on the other hand, determines the set of context words given an input word.

The CBoW and SG models consider words as basic units ignoring rich subword information, thereby, significantly limiting the performance of the models [11]. This shortcoming was addressed in [1] by extending the SG model using subword information. Each word is represented as a bag of character n-grams and the word vector is derived as the summation of the n-gram vectors.

Global Vectors, GloVe [12], is another model for word representation that was proposed to address the drawbacks suffered by CBoW and SG models. The models train on separate local context windows instead of global word-word co-occurrence counts, thereby, capturing and making efficient use of global statistics. It was shown to outperform word2vec [9] on word analogy, named-entity-recognition and word similarity tasks.

**Related work.** Many works have been carried out on generating words embedding for a number of languages. Example of such works includes that of [13] that provides Arabic words embedding for NLP. [14] provides word vectors for 100 languages to be used for multilingual NLP. [15] also provided words embedding for 157 rich and under-resourced languages. None of these works includes Hausa language. The only publicly available work on word vectors representation that includes Hausa language among other 293 languages is that of [1]. The Hausa words embedding consists of about 4k words in Hausa, English and other languages.

The rest of the paper is organized as follows: in section II we describe the methodology which includes data generation, data cleaning and other preprocessing processes, and model generation. In section III, the models are evaluated qualitatively and finally, in section IV, we conclude the paper and discuss future work.

## II. METHODOLOGY

### A. Data Generation and Preprocessing

The models are trained using datasets composing of the Tanzil dataset and crawled news articles from the web.

**Tanzil:** The Tanzil dataset is a translation of the Quran in various languages including Hausa. The monolingual Hausa dataset consists of 127k sentences. The model will be trained on the Hausa monolingual dataset in the Tanzil corpus.

**Web Crawl:** news data collected from Hausa news sites in the world wide web using crawlers created for each news site to maximize the amount of data collected. The crawled news data consists of data in different domains such as sports, literature, finance, education, culture, security, etc.

The statistics of all data used in training the model is showed in Table 1.

The following are the data cleaning and preprocessing processes carried out on the raw data before training

1. Removing punctuations
2. Removing sentences made only of characters, not words
3. Removing numbers and non-*Boko* scripts
4. Minimum words count in a sentence is set to 3
5. Tokenization was done using a python script created for this task
6. Duplicate sentences were removed.

To determine the number of English words in the entire corpus both in the old and new models, each word in the models' vocabularies was checked against the Wordnet dataset. The Wordnet dataset is an English dictionary contained in the Natural Language Toolkit (NLTK) [16]. Some of the words appear both as Hausa and English words. We, therefore, removed some of them which we were sure they are used in the documents as Hausa words. These words are: *a, da, ta, ya, na, ba, yi, su, yi, ne, ce, shi, ga, za, sai, yan, aka, wa, kan, nan, ko, ka, hau, mu, masu, kasa, kai, dan, ake, sa, amma, yana, yin, tare, bai, ita, ni, baya, ana, masa, din, tun, mun, kafa, dama, akan, ji, zaman, fi, tana, zo, abu, kama, mana, sha, kula, zan, jin, kayan, boko, ki, dole, babu, dace, gare, dauke, damar, kansa, kashi, rana, dari*.

Determining the number of English words is to ascertain the usability or otherwise of the model as Hausa words embedding. An embedding that contains the majority of its words vectors as English words would not be appropriate to be used for Hausa NLP tasks.

*B. The Models*

The models were trained using gensim, a tool created by [14] for various NLP tasks such as topic modelling and word vectors generation.

Two models were created: hauWE CBOW and hauWE SG based on the implementation by [1].

The models were trained for 5 epochs setting the minimum word count to 1 and model size as 300. All other parameters were left as default as set in gensim's Word2Vec model implementation. This means an initial learning rate of 0.025 which will, as the training progresses, drop to 0.0001. The default window size is set to 5. The training is done using negative sampling with 5 negatives or "noise words" to be drawn.

The trained models are available for research at https://abumafrim.github.io/main/2019/06/14/hauWE.html.

## III. EVALUATIONS

For evaluating the models, we use the model to generate similar Hausa words. The two models perform better than Bojanowski.

The Hausa words used are: *miji, mata, makaranta, gida, tafiya, kyau, ido, waya, kira, hadisi, kara, godiya, kuka, ibrahim, so, kallo, unguwa, dariya, kaga, sai, kuma, rawa, kida, waka, habiba, zainab, kalma, musa, abdullahi* and *littafi*. These include common day-to-day spoken words and names of individuals.

| words | similar words predicted correctly | | |
|---|---|---|---|
| | Bojanowski | hauWE CBOW | hauWE SG |
| *miji* | 0 | 10 | 10 |
| *mata* | 1 | 10 | 9 |
| *makaranta* | 8 | 10 | 10 |
| *gida* | 3 | 7 | 6 |
| *tafiya* | 2 | 8 | 9 |
| *kyau* | 1 | 9 | 8 |
| *ido* | 0 | 8 | 10 |
| *waya* | 0 | 10 | 10 |
| *kira* | 5 | 10 | 9 |
| *hadisi* | 3 | 10 | 10 |
| *kara* | 2 | 8 | 7 |
| *godiya* | 0 | 10 | 10 |
| *kuka* | 1 | 9 | 4 |
| ibrahim | 5 | 10 | 10 |
| *so* | 0 | 9 | 5 |
| *kallo* | 1 | 7 | 7 |
| *unguwa* | 4 | 10 | 8 |
| *dariya* | 0 | 9 | 8 |
| *kaga* | 1 | 7 | 6 |
| *sai* | 0 | 10 | 0 |
| *kuma* | 0 | 8 | 0 |
| *rawa* | 1 | 3 | 4 |
| *kida* | 0 | 10 | 10 |
| *waka* | 0 | 10 | 10 |
| habiba | 1 | 10 | 10 |
| zainab | 2 | 10 | 10 |
| *kalma* | 0 | 4 | 7 |
| musa | 8 | 10 | 10 |
| abdullahi | 9 | 10 | 10 |
| *littafi* | 9 | 10 | 9 |
| | 22.3% | **88.7%** | 79.3% |

*Table 2 Similar words predicted correctly by the models given an input word. The words predicted are shown in Appendix A*

The model was used to predict 10 most similar words, given an input word. 30 common words and names were used as input and the statistic of the prediction is showed in Table 2. hauWE CBOW produced the best result, predicting 88.7% words accurately whereas Bojanowski performed the worst with just 22.3% accuracy. The huge accuracy margin can be attributed to the difference in the size of vocabulary and the amount of training data.

Both models – hauWE CBOW and hauWE SG – were accurate in predicting only female named entities when a female name was given as input and also, predicted male named entities when a similar word is given as input.

The accuracy of the two models over Bojanowski can also be attributed to the ratio of English to Hausa words in the training corpus. The hauWE models and Bojanowski contain 1:14 and 1:5 English to Hausa words respectively. 38% of the vocabulary in the Bojanowski model are English words which translate to 21% of the entire corpus. The huge amount of English words in Bojanowski influenced the accuracy of the model to learn good vectors for Hausa words. This can be seen in Table 3 where the similar words predicted by Bojanowski are mostly not Hausa words.

## IV. CONCLUSION

In this work, we created word vector models using the CBoW and SG for Hausa NLP research. The set of models, hauWE, consists of a considerably larger amount of words in the vocabulary and has been shown to perform considerably better than the previous publicly available model. hauWE CBoW, in particular, outperforms the two models. This confirms the hypothesis by [8] that for a large dataset, CBoW model outperforms SG model. In future work, we aim to create comprehensive test sets, especially analogy datasets, to be able to measure the accuracy of Hausa words embedding. More text resources will also be generated to improve the performance of the models. Lastly, some tasks in NLP require in-domain text resources. In the future, in-domain data will be used to create words embedding for domain-specific tasks such as in medicine.

## References

[1] P. Bojanowski, E. Grave, A. Joulin, and T. Mikolov, "Enriching Word Vectors with Subword Information," *CoRR*, vol. abs/1607.0, 2016.

[2] D. Bahdanau, K. Cho, and Y. Bengio, "Neural Machine Translation by Jointly Learning to Align and Translate," *arXiv Prepr. arXiv1409.0473*, 2014.

[3] M. Artetxe and H. Schwenk, "Massively Multilingual Sentence Embeddings for Zero-Shot Cross-Lingual Transfer and Beyond," *arXiv:1812.10464v1*, 2018.

[4] M. Giatsoglou, M. G. Vozalis, K. Diamantaras, A. Vakali, G. Sarigiannidis, and K. C. Chatzisavvas, "Sentiment analysis leveraging emotions and word embeddings," *Expert Syst. Appl.*, vol. 69, pp. 214–224, 2017.

[5] P. Wang, B. Xu, J. Xu, G. Tian, C. Liu, and H. Hao, "Semantic expansion using word embedding clustering and convolutional neural network for improving short text classification," *Neurocomputing*, vol. 174, pp. 806–814, 2016.

[6] N. Pourdamghani, M. Ghazvininejad, and K. Knight, "Using Word Vectors to Improve Word Alignments for Low Resource Machine Translation," in *Proceedings of NAACL-HLT*, 2018, pp. 524–528.

[7] P. J. Jaggar, "Hausa," *Elsevier Ltd*, pp. 222–225, 2006.

[8] A. Joulin, E. Grave, P. Bojanowski, M. Douze, H. Jégou, and T. Mikolov, "FastText.zip: Compressing text classification models," *arXiv Prepr. arXiv1612.03651*, 2016.

[9] T. Mikolov, G. Corrado, K. Chen, and J. Dean, "Efficient Estimation of Word Representations in Vector Space," *arXiv:1301.3781v3*, pp. 1–12, 2013.

[10] G. Berardi, A. Esuli, and D. Marcheggiani, "Word Embeddings Go to Italy : a Comparison of Models and Training Datasets," in *Proceedings of the Italian Information Retrieval Workshop*, 2015.

[11] W. Chen and W. Sheng, "A Hybrid Learning Scheme for Chinese Word Embedding," in *Proceedings of the 3rd Workshop on Representation Learning for NLP*, 2018, pp. 84–90.

[12] J. Pennington, R. Socher, and C. D. Manning, "GloVe : Global Vectors for Word Representation," in *Proceedings of the 2014 Conference on Emperical Methods in Natural Language Processing (EMNLP)*, 2014, pp. 1532–1543.

[13] A. B. Soliman, K. Eissa, and S. R. El-Beltagy, "AraVec: A set of Arabic Word Embedding Models for use in Arabic NLP," in *3rd International Conference on Arabic Computational Linguistics*, 2017, vol. 117, pp. 256–265.

[14] R. Al-Rfou, B. Perozzi, and S. Skiena, "Polyglot: Distributed Word Representations for Multilingual NLP," *arXiv:1307.1662v2*, 2014.

[15] E. Grave, P. Bojanowski, P. Gupta, A. Joulin, and T. Mikolov, "Learning Word Vectors for 157 Languages," *arXiv:1802.06893v2*, pp. 1–5, 2018.

[16] S. Bird, L. Edward, and K. Ewan, *Natural Language Processing with Python*. O'Reilly Media Inc., 2009.

# APPENDIX A

| words | models | similar words |
|---|---|---|
| miji | hauWE CBOW | *namiji, mijinta, mijin, kishiya, maigida, matarsa, mahaifi, aure, mahaifiya, macen* |
| | hauWE SG | *namiji, mijinta, mijin, mijinsu, mijinki, auro, manemin, mace, saketa* |
| | Bojanowski | *daidaito, yunwa, daidai, zagaye, firinji, gwargwadon, yiwa, alheri, kura, dawowa* |
| mata | hauWE CBOW | *matan, matansu, marayu, mace, zawarawa, yayansu, mãtã, yara, mazaje, iyaye* |
| | hauWE SG | *matan, maza, damata, yaranta, yanmata, iyaliina, yara, samari, mace, yammata* |
| | Bojanowski | *kamata, fata, matar, huta, zata, sata, cutarwa, nata, fita, ambata* |
| makaranta | hauWE CBOW | *jamia, makarantar, makarantarmu, makarantu, firamare, makarantan, makarantun, makaratar, kwaleji, sakandare* |
| | hauWE SG | *makarantar, makarantan, makarata, makarantarmu, jamia, islamiya, islamiyyar, makaratar, islamiyya, makarantarsu* |
| | Bojanowski | *makarantun, karanta, makarantu, makarantar, karantarwa, karantar, karatunsa, karatun, makkah, burujerdi* |
| gida | hauWE CBOW | *daki, gidanta, gidansu, hayyacinsu, hayyacinta, gidana, shago, hayyacinsa, gidajensu, gidanmu* |
| | hauWE SG | *hayyacinsa, falo, daki, hayyacin, siton, hayyacina, waje, gidansu, kayayyakina, dakina* |
| | Bojanowski | *gidaje, gudu, biyu, gidajen, hudu, uku, biyo, watsi, gidan, gishiri* |
| tafiya | hauWE CBOW | *juyawa, tafiyar, wucewa, zagayawa, yawo, hutawa, taruwa, lalacewa, dawowa, canzawa* |
| | hauWE SG | *tafiyar, tafiyarsu, tafi, tafiyarsa, dakatawa, jiransa, sassarfa, firgice, karasawa, balaguro* |
| | Bojanowski | *tafiyarki, mafiya, lafiya, miliyoyin, gaskiya, kimiya, tasowa, talauci, tsinci, burbushin* |
| kyau | hauWE CBOW | *kyawu, inganci, tsabta, nagarta, maana, muhimmanci, kayatarwa, sauki, tsafta, mahimmanci,* |
| | hauWE SG | *kyawu, inganci, mahimmaci, bullewa, gautsi, araha, kayatarwa, kayu, muhimmanci, rudarwa,* |
| | Bojanowski | *kyauta, gungun, magunguna, abinda, fahimta, kyautata, lalata, koyi, tabbata, taimako* |
| ido | hauWE CBOW | *idanu, idon, klwallaye, idanunsu, idonta, gugansa, hijabi, saido, hanu, idonsu,* |
| | hauWE SG | *idanu, idon, idonta, idanuwansu, idanunsu, saido, idonsu, idanun, rintse, hazohazo,* |
| | Bojanowski | *canji, fahimta, canje, rayuwarsu, koyi, gungun, taimako, magunguna, zuba, tsoro* |
| waya | hauWE CBOW | *wayarsa, lambarsa, tes, wayata, wayarta, wayan, wayarka, wayoyinsu, imel,* blackberry, |
| | hauWE SG | *wayarsa, lambarsa, salula, wayarta, tes, tarho, wayata, gsm, salular, bugo,* |
| | Bojanowski | *mota,* nobel, *wasanni, kwayoyin, dinga, fasahar, wasan, wasa, shayar, samuwar* |
| kira | hauWE CBOW | *kiran, kirana, k̂ira, gargadi, kiranka, shawarci, shawartar, roko, bukaci, godiya,* |
| | hauWE SG | *kiran, k̂ira, tattausar, kirana, bukaci, shawartar, shawararka, kiranka, kiranki, shawarci,* |
| | Bojanowski | *zurfi, karshen, kiransa, kirari, hira, misira, girmansa, suyi, kiransu, sulhu* |
| hadisi | hauWE CBOW | *hadisin,* ibnu, bukhari, *sahihul,* tirmizi, *ibn, hibban, mujalladi, khaldun,* muslim, |
| | hauWE SG | *hadisin, ibnu,* tirmizi, *sahihul,* bukhari, *isnadi, mujalladi, hudud, hibban,* nasai, |
| | Bojanowski | *hawza, hakuri, hajji, hatsi,* hajj, *agha, fikihu, masallatai,* madina, *ruhu* |
| kara | hauWE CBOW | *kara, qara, dada, rage, karawa, nunar, karin, saukaka, sake, karfafa,* |
| | hauWE SG | *dada, k̂ara, qara, karawa, kudurinsu, yakara, karada, akara, karin, shigarmu,* |
| | Bojanowski | *karami, karo, karfi, kokari, k̂ara, kari, kare, sankara, shawara, karya* |
| godiya | hauWE CBOW | *godiyar, godiyarsa, godiyata, jinjina, godewa, gode, godiyarmu, godiyarsu, taaziyya, mubayaa,* |
| | hauWE SG | *godiyata, godiyarmu, godiyar, godiyarsa, jinjina, godiyarta, gode, godewa, godiyarsu, bangajiya,* |
| | Bojanowski | *italiya, libya, najeriya, kambodiya,* kenya, *gabascin, girka, kudanci,* aljeriya, *koriya* |
| kuka | hauWE CBOW | *kun, ku, kika, muka, kunã, kuke, aka, lãrabci, kanku, zã,* |
| | hauWE SG | *kun, karyatãwa, anani, jinku, ku, dakuka, karkiyata, aryatawa, shirinku, tambayarku,* |
| | Bojanowski | *gaggawa, kusurwa, yiwa, kowacce, iyaye, rinka, duka,* jigawa, *jurewa, kaucewa* |
| ibrahim | hauWE CBOW | aliyu, sulaiman, isah, adamu, bashir, salihu, yahaya, ibrahin, hassan, muhammad, |
| | hauWE SG | ibrahin, ibarahim, ibarhim, muhamamd, abdulhamid, mikailu, galadanci, mahammad, idris, abubukar, |
| | Bojanowski | *alaihi,* jawad, *umarni,* islam, ahmadu, *ummu,* zainab, abdullahi, ahmadi, *alkur* |
| so | hauWE CBOW | *son, sonsa, iyawa, nema, kaunar, tunanin, shaawa, bukata, wulakanta, alfahari,* |
| | hauWE SG | *son, sonsu, sonmu, shugabanka, kudurcewa, shizan, shiina, kaunace, dangantuwa, galibinmu,* |
| | Bojanowski | *can, re, st/acre,* arabic, fine, generated, done, compare, most, related |
| kallo | hauWE CBOW | *kallon, kallonsa, dugwidugwi, kallonsu, sara, kalo, wasansa, sauraro, ihu, tauri,* |
| | hauWE SG | *kallon, kallonsa, kallona, kallonsu, kalo, finafinaina, dariyar,* downloading, *lokeshin, finafinanmu,* |
| | Bojanowski | *kalli, kawai, bahaushe, kadai, koyon, kala, kawo, naka, allura,* jigawa |
| unguwa | hauWE CBOW | *anguwa, unguwar, kauye, dakinsa, gunduma, anguwar, unguwan, kauyen, mazaba, rumfa,* |
| | hauWE SG | *anguwa, dagaci, unguwar, magume, unguwarsu, unguwarmu, gunduma, festac, rugar, bukka,* |
| | Bojanowski | *soji, fito, kauri, ɓangare, ura, sarakuna, gona, gagara,* sahara, *gidansa* |
| dariya | hauWE CBOW | *murmushi, ihu, shaawa, mamaki, haushi, shiru, faxa, burge, tsorata, surutu,* |
| | hauWE SG | *dariyar, zolaya, murmushi, kwakwalwata, kyalkyace, zagina, kallona, kalleka, kyalkyale, mufeeda,* |
| | Bojanowski | *tudu, hudu, kwango, uku, wato, goma, najeriya,* pakistan, *takwas, tura* |
| kaga | hauWE CBOW | *kace, naga, ance, nasan, gashi, akace, suce, wallahi, ai, kam,* |
| | hauWE SG | *zakaga, anzo, suce, zakaji, baayi, akeso, inka, nakeyi, zaice, tinanin,* |
| | Bojanowski | *kenan, zakaga, kalla, habaka,* zidane, *kaka, awannan, yanzu, wannan, akalla* |
| sai | hauWE CBOW | *in, idan, tunda, ashe, bari, gara, shiru, saannan, to, zarar,* |
| | hauWE SG | *wartsake, sia, matsu, hayyacina, hakura, loton, jakata, basai, sannusannu, yanzuyanzu,* |
| | Bojanowski | *sauka, saidai, sako, salla, sace, ubangiji, matuka, zuba, kuwa, haihuwa* |
| kuma | hauWE CBOW | *saboda, domin, da, nufin, don, kuwa, amma, babancin, tare, dai,* |
| | hauWE SG | *jiinci, akenutsarwa, mā, daallah, laifidaga, fakuwa, roi, nasĩha, iblĩs, bayyanannen,* |
| | Bojanowski | *miyagun, kulawa, kula, hukunta, kunun, yazama, tsayawa, kasancewa, kanta, kun* |
| rawa | hauWE CBOW | *tsantsan, rawar, tsan, rawarsu, mahimmiyar, birki, muhimmiyar, burki, takawa, kahankadaran,* |
| | hauWE SG | *taka, muhimmiyar, rawar, mahimmiyar, muhimmiyyar, takawa, muhimiyar, bazarsu, mihimmiyar, rawan,* |
| | Bojanowski | *kulawa, talakawa, awa, sunayen, kwana, kwashe, tsayawa, nawa, kwanan, awannan* |
| kida | hauWE CBOW | *kidan, wakar, rera, waka, waƙa, kidi, wakokin, kadekade, wakokinsa, wakewake,* |
| | hauWE SG | *kidan, kidi, kidansa, kalangu, lela, waka, gurmi, wakokinsa, bakandamiya, wakar,* |
| | Bojanowski | *kiwo, tushe, aji, kanfanoni, gaji, katako, bude, kishirwa, kirkiro, gane* |
| waka | hauWE CBOW | *wakar, wakokin, kida, wakoki, waƙa, wakata, rubutu, rubucerubuce, shata, kidan,* |
| | hauWE SG | *wakar, wakata, wakokin, wakarsa, wakoki, waƙa, wakokinsa, kidan, bakandamiya, kida,* |
| | Bojanowski | *wakanda, kaka, kaucewa, zaka, kotunan, habaka, matuka, wucewa, mummunan, wacce* |
| habiba | hauWE CBOW | bilkisu, mariya, jamila, ummi, hajara, halima, maimuna, hassana, salma, maryam, |
| | hauWE SG | zarah, nusaiba, uwale, faiza, zuwaira, siyama, bilkisu, kanwarta, salma, delu, |
| | Bojanowski | ataturk, *nemo, tsoro, gobe, hakuri, bukatun, waxanda, budurwa, gaggawa, oi* |
| zainab | hauWE CBOW | fatima, amina, hadiza, aishatu, saadatu, hafsat, jummai, rukayya, maryam, jamila, |
| | hauWE SG | fatima, kande, hajiya, babaji, jummai, hafsat, talatu, hajia, bilkisu, aishatu, |
| | Bojanowski | *marmaro,* mirza, *chadi, ppp,* kebbi, *almasihu,* nuruddeen, islam, ummu, jawad |
| kalma | hauWE CBOW | *siffa, manhaja, itaciya, muujiza, taska, masarrafa, kalmar, dabia, maadana, mukala,* |
| | hauWE SG | *maanarta, kalmar, muujiza, taska, siffa, decryption, jimla, kalmomin, kalmomi, asalinta,* |
| | Bojanowski | *kida, gano, matsaloli, kiwo, kirkiro, kanfanoni, gane, kunshi, nasu, akafi* |
| musa | hauWE CBOW | sulaiman, shuaibu, tijjani, rabiu, abdulkadir, jamilu, saidu, muazu, salisu, iliyasu, |
| | hauWE SG | rabiu, isyaku, ismaila, sulaiman, sambawa, auwalu, msa, shamsudden, dangwani, abullahi, |
| | Bojanowski | mustafa, ahmadi, ahmad, khamene, umaru, ahmadu, musulme, jawad, iran, mali |
| abdullahi | hauWE CBOW | umar, sulaiman, jamilu, yahaya, aliyu, muhammad, kabir, haruna, saidu, shuaibu, |
| | hauWE SG | abdulahi, umar, abudllahi, abdulalhi, abudullahi, abdulhamid, ismail, ayagi, abullahi, abdillahi, |
| | Bojanowski | husaini, abdu, sayyid, ahmad, dr, ibrahim, ayatullah, khamene, malam, jawad |
| littafi | hauWE CBOW | *littafin, littafinsa, alurani, rubutu, littattafai, littãfi, attaura, littafina, makala, littattafan,* |
| | hauWE SG | *littafin, juzui, littattafan, littafinsa, littafina, littattafai, littattafansa, tatsuniyoyi, tatsuniyoyin, galatiyawa,* |
| | Bojanowski | *littafin, littafai, littafinsa, littafan, littãfi, littatafai, littattafai, littattafan, sani, li* |

*Table 3 Similar words predicted by the three models*